\pdfoutput=1

\documentclass[11pt]{article}

\usepackage{EMNLP2022}

\usepackage{times}
\usepackage{latexsym}

\usepackage[T1]{fontenc}

\usepackage[utf8]{inputenc}

\usepackage{microtype}

\usepackage{inconsolata}

\usepackage{booktabs}
\usepackage{graphicx}
\usepackage{xcolor}
\usepackage{subcaption}
\usepackage{multirow}
\usepackage{amsmath,amssymb}
\usepackage{adjustbox}
\usepackage{fdsymbol}

\title{FineD-Eval: Fine-grained Automatic Dialogue-Level Evaluation}

\author{Chen Zhang$^{\dag,\star}$ \quad Luis Fernando D’Haro$^\ddag$ \quad Qiquan Zhang$^\dag$ \\ \quad \textbf{Thomas Friedrichs}$^\star$ \quad \textbf{Haizhou Li}$^{\varheartsuit, \dag, \diamondsuit}$     \\
  $^\dag$National University of Singapore \quad $^\star$Robert Bosch (SEA), Singapore \\
  $^\ddag$Universidad Politécnica de Madrid, Spain \quad $^{\diamondsuit}$Kriston AI Lab, China \\
  \quad $^{\varheartsuit}$The Chinese University of Hong Kong, Shenzhen, China \\
        \tt chen\_zhang@u.nus.edu
  }

\begin{document}
\maketitle

\begin{abstract}

Recent model-based reference-free metrics for open-domain dialogue evaluation exhibit promising correlations with human judgment\footnote{As shown in~\citep{yeh-etal-2021-comprehensive}, most reference-free metrics can achieve around 0.3 to 0.6 Spearman correlations on various turn-level benchmarks. However, on the dialogue-level benchmarks, most metrics perform poorly ($<0.2$ Spearman correlations).}. However, they either perform turn-level evaluation or look at a single dialogue quality dimension. One would expect a good evaluation metric to assess multiple quality dimensions at the dialogue level. To this end, we are motivated to propose a multi-dimensional dialogue-level metric, which consists of three sub-metrics with each targeting a specific dimension. The sub-metrics are trained with novel self-supervised objectives and exhibit strong correlations with human judgment for their respective dimensions. Moreover, we explore two approaches to combine the sub-metrics: metric ensemble and multitask learning. Both approaches yield a holistic metric that significantly outperforms individual sub-metrics. Compared to the existing state-of-the-art metric, the combined metrics achieve around 16\% relative improvement 
on average across three high-quality dialogue-level evaluation benchmarks.

\end{abstract}

\section{Introduction}
\label{sec:intro}

In the study of generative dialogue systems, we heavily rely on some reference-based static metrics, such as BLEU~\citep{papineni-etal-2002-bleu}, to measure improvements during system development and to compare various model variants. These metrics still need improvement due to their poor correlations with human judgment~\citep{liu-etal-2016-evaluate} and poor interpretability~\citep{mehri-eskenazi-2020-usr}. 

Recently, model-based reference-free metrics~\citep{yeh-etal-2021-comprehensive} represent one of the ways to address the limitations of static reference-based metrics. Although such metrics exhibit promising correlations with human evaluation, most of them~\citep{tao2018ruber,ghazarian-etal-2019-better,huang-etal-2020-grade,sinha-etal-2020-learning,mehri-eskenazi-2020-usr,phy-etal-2020-deconstruct,pang-etal-2020-towards,zhang-etal-2021-dscore} target turn-level evaluation, i.e., they focus on single-response quality, such as contextual relevance and naturalness. When evaluating a multi-turn human-chatbot dialogue, turn-level metrics do not model the dialogue in totality, but frame it as a set of context-response pairs. They assign scores to every chatbot responses in the dialogue. Hence, an aggregation strategy is required to derive the single dialogue-level metric score, such as taking the average of all the response-level scores. Both prior works~\citep{zhang-etal-2021-dynaeval,yeh-etal-2021-comprehensive} and our experimental results in \S\ref{sec:exp-analysis} suggest that such an approach yields sub-optimal dialogue-level evaluation. The reason may be that turn-level metrics do not model the dependency among utterances within multi-turn interactions, it is difficult for them to spot errors that are only obvious after observing the entire conversation~\citep{NEURIPS2019_fc981212,ghazarian-etal-2022-deam}. 

There are some metrics that perform multi-turn evaluation. However, they focus only on a single dimension, such as coherence or overall impression~\citep{mesgar-etal-2020-dialogue,zhang-etal-2021-dynaeval,li-etal-2021-conversations,ghazarian-etal-2022-deam}. When evaluating a dialogue, they assign a single score to quantify one aspect of dialogue quality. 
As pointed out in~\citet{mehri-etal-2022-nsf}, dialogue quality is inherently multi-faceted. By breaking down the quality of the dialogue into multiple fine-grained dimensions, we may provide a more interpretable and descriptive dialogue evaluation. 
With such an interpretable metric, dialogue researchers know exactly which aspect of the dialogue system to improve.

To this end, we propose a multi-dimensional metric, dubbed FineD-Eval\footnote{\url{https://github.com/e0397123/FineD-Eval}}, which consists of specialized sub-metrics. Each sub-metric targets a specific fine-grained dimension and all sub-metrics are trained in a self-supervised manner without reliance on any human annotations. 


To develop FineD-Eval, our first step is to identify the dimensions for metric design. It is a well-known phenomenon that human judges do not provide completely independent assessments for various fine-grained dimensions. For instance,~\citet{sai-etal-2021-perturbation} analyzes the human ratings with respect to (w.r.t.) different fine-grained dimensions on four text generation tasks and has observed moderate correlations for most dimension pairs. Intuitively, we want to select dimensions that are less correlated such that our metric can holistically capture the dialogue quality from different perspectives. The selection process is guided by an analysis on fine-grained human ratings of dialogue-level evaluation data (\S\ref{sec:human-eval-analysis}). Through the analysis, we want to cluster the dimensions into relatively independent dimension groups and then, select representative dimensions from different dimension groups. 

Next, we propose dimension-specific strategies for training the sub-metrics. (\S\ref{subsec:submetrics}). The sub-metrics, which target the representative dimensions, can also be applied to evaluate other dimensions in their respective dimension groups. Furthermore,  both~\citet{yeh-etal-2021-comprehensive} and~\citet{zhang-etal-2021-automatic} highlight that the combination of different metrics leads to better correlations with human evaluation than individual specialized metrics. We are motivated to explore how to combine the sub-metrics into a unified one. Specifically, both the metric ensemble and multitask learning~\citep{caruana1997multitask} are examined (\S\ref{subsec:combine}). 

Finally, in the experiments (\S\ref{sec:exp-analysis}), we demonstrate that (1) the sub-metrics highly correlate with human judgment for their target dimensions. (2) The scores assigned by FineD-Eval are more interpretable than the existing metrics. (3) With either metric ensemble or multitask learning, FineD-Eval significantly outperforms existing state-of-the-art metrics as well as individual sub-metrics on three high-quality dialogue-level evaluation benchmarks. 

\section{Analysis of Human Evaluation Data}
\label{sec:human-eval-analysis}

\begin{figure}
\centering
\includegraphics[width=\linewidth]{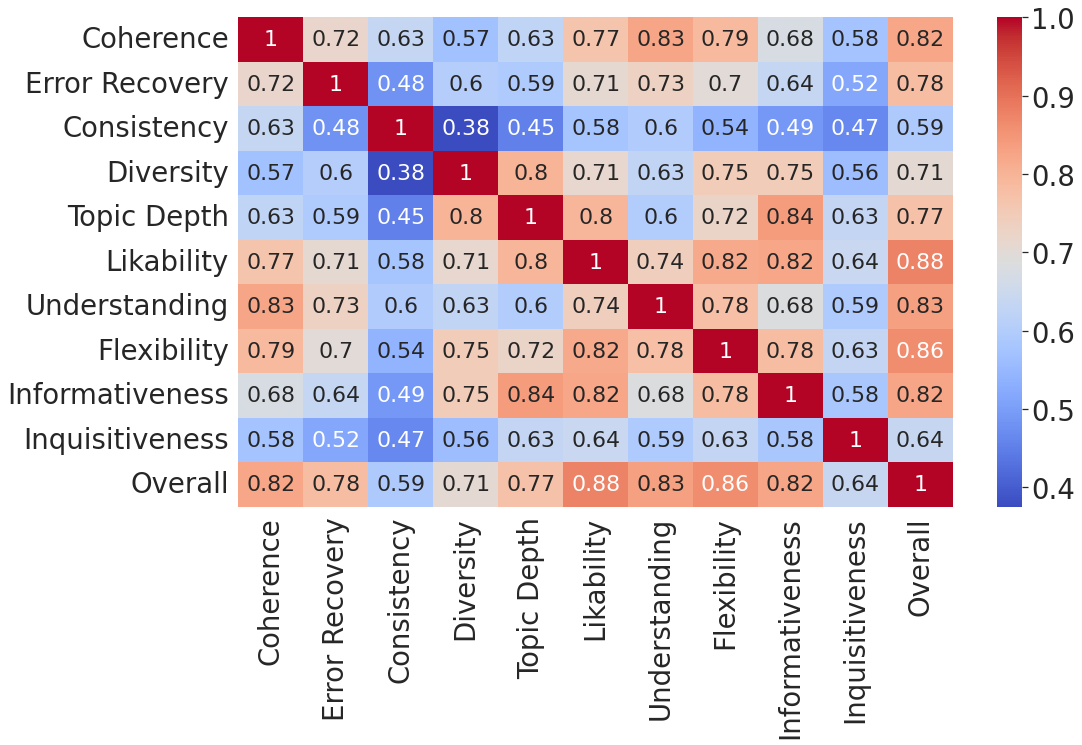}
\caption{Spearman correlations of dimension pairs on FED.}
\label{fig:inter-corr}
\end{figure}

\begin{table}[!t]
\centering
\begin{adjustbox}{max width=0.9\linewidth}
\begin{tabular}{c|c}
\toprule
\textbf{Group} & \textbf{Quality Dimensions} \\ \midrule
Coh & Coherence, Understanding \\
Lik & Likability, Flexibility, Informativeness  \\
Top& Topic Depth, Diversity, Informativeness  \\
Con & Consistency  \\
Inq & Inquisitiveness  \\
Err & Error Recovery \\
\bottomrule
\end{tabular}
\end{adjustbox}
\caption{Grouping of the dimensions. We adopt the first three letters of the representative dimension within each group as the corresponding group name.}
\label{tab:dim-categorization}
\end{table}

\subsection{Grouping of the Dimensions}
\label{subsec:metric-group}

In this section, we analyze the human ratings of FED~\citep{mehri-eskenazi-2020-unsupervised}, a high-quality dialogue-level evaluation benchmark. Each dialogue in FED is annotated by five human judges for 11 different quality dimensions\footnote{The detailed definitions of all dimensions are presented in Table~\ref{tab:dialog-dims} of the Appendix}, as shown in the axis labels of Figure~\ref{fig:inter-corr}. We choose FED for our analysis because the dataset covers the most comprehensive list of dialogue quality dimensions. In addition, the human annotation quality of FED is high as evidenced by the strong inter-annotator agreements w.r.t. different dimensions\footnote{Above 0.75 in terms of Spearman correlations for all the dimensions except that of consistency, which is 0.562.}.

Figure~\ref{fig:inter-corr} presents the Spearman correlations of different dimension pairs on FED. We can observe that all dimensions are interdependent, with correlations ranging from 0.38 to 0.88. Based on their extent of interdependence, we cluster the 10 dimensions (excluding the "Overall" category) into six groups, as shown in Table~\ref{tab:dim-categorization}. We adopt the first three letters of the representative dimension within each group as the corresponding group name. The representative dimension in each group is chosen based on criteria discussed in \S\ref{subsec:metric-select}.

A dimension is treated as an independent group if it does not correlate strongly with any of the other dimensions ($\geq0.75$). Hence, consistency, inquisitiveness, and error recovery can be perceived as three independent dimension groups: \textit{Con}, \textit{Inq}, and \textit{Err} respectively. The remaining dimensions are more or less correlated with each other. Based on the following four observations: (1) coherence strongly correlates with understanding (0.83); (2) The likability-flexibility and likability-informativeness correlations are both 0.82; (3) The correlation between topic depth and informativeness is as high as 0.84; and (4) Diversity only strongly correlates with topic depth (0.8), the remaining seven dimensions can be clustered into three groups: \textit{Coh}, \textit{Lik}, and \textit{Top}.

The categorization may not be perfect as \textit{Coh}, \textit{Lik}, and \textit{Top} are not completely independent from each other. For example, informativeness can be found in both group \textit{Lik} and group \textit{Top}. A possible explanation is that humans generally like knowledgeable chatbots, which can discuss different topics in depth rather than those that generate dull responses~\citep{see-etal-2019-makes,roller-etal-2021-recipes}. To improve the categorization, future work may conduct similar analysis on large-scale dialogue-level human annotations.

\subsection{Dimension Selection}
\label{subsec:metric-select}

As metioned in \S\ref{sec:intro}, we want to identify fine-grained dimensions that are less similar. Hence, we select only one dimension from each group and avoid those that are shared between two different groups. In addition, to further reduce the complexity of FineD-Eval, we implement the following rules to narrow down the selection to only three fine-grained dimensions. 

First, dimensions that highly correlate with the "Overall" category (> 0.75) are considered. The intuition is that a high correlation with "Overall" indicates more influence from the fine-grained dimension on human annotators' overall impression about a dialogue. Second, we filter out dimensions with low inter-annotator agreement (< 0.6)\footnote{Only consistency has an inter-anntator agreement below 0.6.}, because low inter-annotator agreements may suggest the dimension is complex to evaluate and human annotators have different understandings of the dimension~\citep{mehri-etal-2022-nsf}. Lastly, we choose dimensions based on how often they are marked as "N/A" (not applicable) by the human judges. A high frequency indicates that the dimension is not generally applicable in different contexts. Most dimensions do not contain a "N/A" rating except "Error recovery", which has been marked as "N/A" 25\% of the time.



Based on the rules, we choose the following three dimensions: \textbf{coherence}, \textbf{likability}, and \textbf{topic depth}. In addition to the rules, we choose these dimensions because they are also widely studied in open-domain dialogue systems. Researchers spend significant amount of efforts on developing coherent, engaging, and knowledgeable chatbots~\citep{adiwardana2020towards,hedayatnia-etal-2020-policy,shuster-etal-2021-retrieval-augmentation}. Designing meaningful metrics along these three dimensions can benefit the current open-domain dialogue research. Though other dimensions, such as consistency~\cite{nie-etal-2021-like}, inquisitiveness~\cite{see-etal-2019-makes}, and long-term memory~\citep{xu-etal-2022-beyond} are equally important, their evaluation deserves a thorough study on its own. Hence, we leave them for future work.

\section{Methodology}
\label{sec:methodology}

\subsection{Problem Formulation}
\label{subsec:formulation}


We formally define the dialogue-level evaluation task. Suppose that we have a dialogue evaluation dataset, $\mathbb{D}$, which contains $n$ human-chatbot dialogues,  $\mathbb{D} = \{d_1, d_2, \ldots, d_j, \ldots, d_{n}\}$. $d_j$ is annotated by several human judges for a set of quality dimensions, $\mathbb{Q}$. Each human judge provides a rating to $d_j$ for individual dimension, $q\in{\mathbb{Q}}$. We use $r_{d_j}^q$ to denote the average Likert rating provided by all human annotators to $d_j$ for $q$. 

Our goal is to learn dimension-specific metrics, $M^q(d_j) \rightarrow s_{d_j}^q$, where $s_{d_j}^q$ is the metric score reflecting how good $d_j$ is for dimension $q$ as perceived by $M^q$. To assess the performance of $M^q$ on $\mathbb{D}$, the correlation, denoted as $\rho^q$, between $S^q = \{s_{d_1}^q, \ldots, s_{d_j}^q, \ldots, s_{d_n}^q\}$ and $R^q = \{r_{d_1}^q, \ldots, r_{d_j}^q, \ldots, r_{d_n}^q\}$ are calculated. Higher $\rho^q$ indicates better performance of $M^q$ on $\mathbb{D}$.


\subsection{General Framework}
\label{subsec:general-framework}

We propose a multi-dimensional dialogue-level metric, FineD-Eval, which is a combination of three specialized sub-metrics, $M^q$, where $q\in \{\text{coherence}, \text{likability}, \text{topic depth}\}$. We explore two approaches for combining the sub-metrics, metric ensemble and multitask learning. Metric ensemble is a late fusion approach whereby the predictions made by the sub-metrics are combined. Multitask learning, on the other hand, is an early fusion approach whereby the sub-metrics will share a common text encoder while having different output layers. Details of both approaches are discussed in \S\ref{subsec:combine}. Here, we focus on the details of $M^q$. 

To train $M^q$, we formulate a preference learning approach~\citep{furnkranz-etal-2011-preference}. Given a pair of dimensions-specific positive and negative training dialogue samples, $d_{tr}^{+}$ than $d_{tr}^{-}$,  $M^q$ learns to predict a higher score for $d_{tr}^{+}$ than $d_{tr}^{-}$. The strategies for constructing ($d_{tr}^{+}$, $d_{tr}^{-}$) are outlined in~\S\ref{subsec:submetrics}. During training, a mini-batch is formed with two types of data instances\footnote{This formulation is to avoid model relying on positions of the dialogues to make predictions.}: (1) ($d_{tr}^{+}$, $d_{tr}^{-}$) with label $y = 1$; (2) ($d_{tr}^{-}$, $d_{tr}^{+}$) with label $y = -1$. $M^q$ outputs two scalar values $s_{d_{tr}^{+}}^q$ and $s_{d_{tr}^{-}}^q$ that correspond to $d_{tr}^{+}$ and $d_{tr}^{-}$ respectively. The following margin ranking loss is adopted to train the model: 
\begin{equation}
 \mathcal{L}_q = max(0, y * (x^q_1 - x^q_2) + 0.1)  
\end{equation}
where $(x_1^q, x_2^q, y)$ can be either ($s_{d_{tr}^{+}}^q$, $s_{d_{tr}^{-}}^q$, 1) or ($s_{d_{tr}^{-}}^q$, $s_{d_{tr}^{+}}^q$, -1). 

The pairwise ranking formulation is motivated by previous works on dialogue evaluation~\citep{mesgar-etal-2020-dialogue,huang-etal-2020-grade,gao-etal-2020-dialogue,zhang-etal-2021-dynaeval}. Compared to direct assessment approaches~\citep{zhang-etal-2021-dscore,ghazarian-etal-2022-deam}, the main advantage of pairwise ranking is that the model can implicitly learn the features that distinguish the good dialogues from the bad ones based on a large quantity of dialogue pairs for a specific quality dimension.

The network architecture of $M^q$ is straightforward. RoBERTa-base~\citep{liu2019roberta} is adopted as the text encoder, $\mathcal{T}$, which maps ($d_{tr}^{+}$, $d_{tr}^{-}$) to dense representations ($\mathbf{H}_{tr}^{+}$, $\mathbf{H}_{tr}^{-}$). Both $d_{tr}^{+}$ and $d_{tr}^{-}$ are formulated as a token sequence with special token "</UTT>" to delimit different utterances. Next, ($\mathbf{H}_{tr}^{+}$, $\mathbf{H}_{tr}^{-}$) are converted into vector representations ($\mathbf{h}_{tr}^{+}$, $\mathbf{h}_{tr}^{-}$) with average pooling. Through a linear layer with output size 1 and a Sigmoid activation function, $\mathbf{h}_{tr}^{+}$ and $\mathbf{h}_{tr}^{-}$ are transformed into scalar values $s_{d_{tr}^{+}}^q$ and $s_{d_{tr}^{-}}^q$ respectively. During inference, given $d_j\in{\mathbb{D}}$, the scalar value $s_{d_j}^q$ output by $M^q$ is the corresponding metric score.

\subsection{Dimension-Specific Sampling Strategies}
\label{subsec:submetrics}

In this section, we discuss different strategies to obtain dimension-specific training dialogue pairs. All ($d_{tr}^{+}$, $d_{tr}^{-}$) samples are automatically constructed from human-human dialogue datasets without reliance on human annotations.

\bigskip
\noindent \textbf{Coherence (\textit{Coh})}\ We consider two strategies for coherence. The first is utterance order shuffling whereby dialogues from existing human-human dialogue corpora~\citep{li-etal-2017-dailydialog,dinan2020second} are treated as $d_{tr}^{+}$. To obtain $d_{tr}^{-}$, we randomly permute the order of utterances in $d_{tr}^{+}$. This strategy has been widely adopted in previous dialogue coherence studies~\citep{cervone18_interspeech,mesgar-etal-2020-dialogue,zhang-etal-2021-dynaeval}.

The second strategy, question-answer (QA) relevance scoring, is motivated by the Gricean maxims~\citep{grice1975logic} whereby effective communication involves being relevant, i.e., one should provide information that is relevant to the current exchange. A natural and logical flow of conversation often involves asking and answering questions, which is a form of information exchange. Humans usually prefer answers that are straight to the point rather than those that are vague and off-topic. Concretely, we select dialogues in existing dialogue corpora\footnote{We hypothesize that even in human-human dialogue corpora, there are answers that are vague and off-topic due to the presence of low-quality crowd-source workers.} that are more than 4 utterances and contain at least one question-answer pair. Next, we use a pretrained BERT-based QA evaluator from HuggingFace\footnote{\url{https://huggingface.co/iarfmoose/bert-base-cased-qa-evaluator}} to score each QA pair within a dialogue. The evaluator provides a relevance score between 0 and 1 (the higher the better). Then, we average the relevance scores of all QA pairs within the dialogue to derive the dialogue-level QA relevance score. Finally, two thresholds, ($\tau^{rel}_{low}$, $\tau^{rel}_{high}$), are chosen. Dialogues with scores lower than $\tau^{rel}_{low}$ are considered $d_{tr}^{-}$. Those with scores higher than $\tau^{rel}_{high}$ are considered $d_{tr}^{+}$. ($\tau^{rel}_{low}$, $\tau^{rel}_{high}$) are heuristically determined to ensure sufficient data in both the positive and negative classes. 

\bigskip
\noindent \textbf{Likability (\textit{Lik})}\ Two strategies are applied to construct $d_{tr}^{+}$ and $d_{tr}^{-}$ for likability. The first strategy, contradiction scoring, is motivated by the similartity attaction effect~\citep{byrne1968effects,nass2001does}. During human-human interaction, people tend to favour others who share similar opinions or preferences with them. On the contrary, conveying contradictory opinions or information may lead to disagreement and user dissatisfaction. 

To implement this strategy, we adopt a pre-trained natural language inference (NLI) model\footnote{\url{https://huggingface.co/roberta-large-mnli}.} to provide contradiction scores (between 0 and 1) to adjacent utterance pairs within human-human dialogues. For a dialogue containing $k$ utterances, we have $k-1$ adjacency pairs, thus $k-1$ contradiction scores. The dialogue-level contradiction score is derived by computing the average of the $k-1$ scores. Finally, two thresholds, ($\tau^{contra}_{low}$, $\tau^{contra}_{high}$), are set. Dialogues with contradiction scores lower than $\tau^{contra}_{low}$ are considered $d_{tr}^{+}$ and those with scores higher than $\tau^{contra}_{high}$ are considered $d_{tr}^{-}$.

The second strategy is based on the number of utterances that carry positive emotions within a dialogue, which we hypothesize can serve as a proxy indicator on how much the interlocutors enjoy conversing with each other. Intuitively, if a user feels a dialogue system is likeable, they tend to produce more engaging responses. To implement the strategy, we adopt a pre-trained sentiment classification model\footnote{\url{https://huggingface.co/mrm8488/t5-base-finetuned-emotion}} and apply it to classify the sentiments w.r.t. all utterances within a dialogue. We treat dialogues, of which all utterances are classified into the positive classes, as $d_{tr}^{+}$ and those containing less than two positive utterances as $d_{tr}^{-}$. 

\bigskip
\noindent \textbf{Topic Depth (\textit{Top})}\ \ Discussing topics in depth is an important attribute of engaging conversations. During the human-human interaction, when the interlocutors deeply dive into a topic, they tend to produce semantically diverse utterances, which convey a large amount of information. On the other hand, if an interlocutor is not interested in the topic, they tend to produce dull responses, such as "Ok", "Good to know", and "I don't know". Even though, such responses can be appropriate in a wide range of contexts, they often do not convey much information~\citep{see-etal-2019-makes}. As most human-human dialogues are topic coherent, we can directly link topic depth to how semantically different the utterances are within a dialogue. Hence, we propose an entailment scoring strategy. 

More specifically, given a dialogue of $k$ utterances, a pre-trained NLI model\footnote{Same as that used for the contradiction scoring strategy} is used to provided entailment score to each utterance pair in the dialogue. In total, there are $\frac{(k-1)k}{2}$ entailment scores per dialogue. The dialogue-level entailment score is the average of all utterance-pair entailment scores in the dialogue. Similarly, two thresholds, ($\tau^{entail}_{low}$, $\tau^{entail}_{high}$), are applied to obtain positive and negative dialogues. Dialogues with entailment scores lower than $\tau^{entail}_{low}$ are regarded as $d_{tr}^{+}$ and those with scores higher than $\tau^{entail}_{high}$ are $d_{tr}^{-}$.

\subsection{Combining Dimension-Specific Metrics}
\label{subsec:combine}

Our analysis in \S\ref{sec:human-eval-analysis} suggests that human evaluation across different quality dimensions are positively correlated. Therefore, a sub-metric that is specialized in evaluating one dimension can contribute to the evaluation of other dimensions as well. By combining different sub-metrics into a holistic one, we can achieve better correlations with human evaluation across different dimensions. We implement two FineD-Eval variants, $\text{FineD-Eval}_{en}$ (metric ensemble) and $\text{FineD-Eval}_{mu}$ (multitask learning).


\bigskip
\noindent \textbf{Metric Ensemble} \ Ensemble is a common technique adopted in machine learning to achieve better predictive performance than individual predictive models. In addition, it also helps improve model robustness by reducing the spread or dispersion of the predictions~\citep{zhang2012ensemble}. 

In our case, $\text{FineD-Eval}_{en}$ is expected to achieve better $\rho^q$ than $M^q$ on $\mathbb{D}$. Given $d_j\in{\mathbb{D}}$, three sub-metrics, $M^{\text{coh}}$, $M^{\text{lik}}$, and $M^{\text{top}}$ output three scores, $s_{d_j}^{\text{coh}}$, $s_{d_j}^{\text{lik}}$, and $s_{d_j}^{\text{top}}$ respectively. The metric score of $\text{FineD-Eval}_{en}$, $s_{d_j}^{\text{en}}$ is obtained by computing the arithmetic mean of ($s_{d_j}^{\text{coh}}$, $s_{d_j}^{\text{lik}}$, $s_{d_j}^{\text{top}}$).  

\bigskip
\noindent \textbf{Multitask Learning} \ In multitask learning, a model is trained simultaneously with multiple tasks and a shared representation is learned to capture the commonalities among the related tasks~\citep{crawshaw2020multi,gao2021tran, chen-etal-2021-revisiting}. Compared to $\text{FineD-Eval}_{en}$, the multitask model, $\text{FineD-Eval}_{mu}$, requires much less model parameters, but can achieve similar performance. 


Similarly, $\text{FineD-Eval}_{mu}$ is also expected to achieve better $\rho^q$ than $M^q$ on $\mathbb{D}$. To implement $\text{FineD-Eval}_{mu}$, we need to first identify the related tasks for joint training. As described in \S\ref{subsec:general-framework}, we have the preference learning tasks for $M^{\text{coh}}$, $M^{\text{lik}}$, and $M^{\text{top}}$ respectively. Since the input and output of the three tasks are the same, we can adopt a hard-parameter sharing network to simultaneously learn the three tasks. More specifically, the text encoder $\mathcal{T}$, is shared among the three tasks. On top of $\mathcal{T}$, there are three independent linear layers with output size 1, which serve as the sub-metrics for coherence, likability, and topic depth respectively. 

During training, a mini-batch consists data that are uniformly drawn from the three training data sources described in \S\ref{subsec:submetrics}. The parameter update of $\mathcal{T}$ depends on all data instances in the mini-batch while that of the three linear layers depends only on their corresponding task-specific input data. The losses of three tasks are summed together, $\mathcal{L}_{total} = \mathcal{L}_{coh} + \mathcal{L}_{lik} + \mathcal{L}_{top}$.

During inference, given $d_j\in{\mathbb{D}}$, $\text{FineD-Eval}_{mu}$ outputs three scalar values, $s_{d_j}^{\text{coh}}$, $s_{d_j}^{\text{lik}}$, and $s_{d_j}^{\text{top}}$ from the three linear layers respectively. Similar to metric ensemble, the final metric score, $s_{d_j}^{mu}$ is derived by taking the arithmetic mean of the three scores.


\section{Experimental Setup}
\label{sec:setup}

\subsection{Training \& Evaluation Datasets}
\label{subsec:train-eval-data}

For training, we prepare two datasets leveraging DailyDialog (DD)~\citep{li-etal-2017-dailydialog} and ConvAI2 (CA)~\citep{dinan2020second}. DailyDialog covers general day-to-day topics, such as school, work, and relationship. ConvAI2 is an extended version of PersonaChat~\citep{zhang-etal-2018-personalizing}, which contains dialogues grounded by persona profiles. The detailed descriptions of DailyDialog and ConvAI2 are included in Appendix~\ref{sec:detail-dataset}. We choose DailyDialog and ConvAI2 because they cover a diverse sets of topics and our baseline metrics (\S\ref{subsec:baseline-metrics}) are mainly trained with these two datasets. The numbers of $d_{tr}^{+}$ and $d_{tr}^{-}$ obtained with various sampling strategies (\S\ref{subsec:submetrics}) are listed in Table~\ref{tab:train-statistics}. When training each $M^q$ on each dataset, we sample 100K and 10K of training and validation dialogue pairs respectively due to the large number of ($d_{tr}^{+}$, $d_{tr}^{-}$) combinations.

\begin{table}[!ht]
\centering
\begin{adjustbox}{max width=0.9\linewidth}
\begin{tabular}{l|ccc}
\toprule
\textbf{Attributes} & \textbf{FED} & \textbf{DSTC9} & \textbf{P-Eval} \\ \midrule
\#Dialogues & 125 & 2,200  & 3,316 \\
Avg. \#Utts per Dialogue & 12.72 & 28.13 & 16.04 \\
Avg. \#Words per Utt & 8.95 & 8.58 & 5.68 \\
\#Dimensions & 11 & 11 & 8\\
\#Ratings & 23,750 & 71,203 & 26,528 \\
\#Models & 3 & 10 & 29 \\
\bottomrule
\end{tabular}
\end{adjustbox}
\caption{Statistics of the three evaluation benchmarks. "P-Eval" refers to the Persona-Eval benchmark.}
\label{tab:eval-statistics}
\end{table}

Three benchmarks are adopted to assess the strength of the metrics. They are FED~\citep{mehri-eskenazi-2020-unsupervised}, DSTC9-Interactive~\citep{gunasekara-etal-2020-dstc9}, and Persona-Eval~\citep{see-etal-2019-makes}. The benchmarks' statistics are shown in Table~\ref{tab:eval-statistics} and their descriptions are presented in Appendix~\ref{sec:detail-benchmark}. The definitions of various quality dimensions of the benchmarks are listed in Table~\ref{tab:dialog-dims} and Table~\ref{tab:persona-eval-dims}. All metrics are assessed with dialogue-level Spearman correlations w.r.t. each fine-grained dimension on the three benchmarks. Note that we do not consider inquisitiveness, consistency, and error recovery in the main analysis, because none of the FineD-Eval sub-metrics target these dimensions. Nevertheless, we show the metrics' performance for the three dimensions in the Limitation section.



\subsection{Baselines}
\label{subsec:baseline-metrics}

Two groups of metrics are adopted. The first are state-of-the-art turn-level metrics, including USL-H~\citep{phy-etal-2020-deconstruct}, MAUDE~\citep{sinha-etal-2020-learning},  MDD-Eval~\citep{zhang2021mdd}, and D-score~\citep{zhang-etal-2021-dscore}. Turn-level metrics need to rely on aggregation strategies when evaluating multi-turn dialogues. In this paper, we adopt mean aggregation whereby the metric scores w.r.t. all chatbot turns in a dialogue are averaged to derive the single dialogue-level metric score. The second group includes  DynaEval~\citep{zhang-etal-2021-dynaeval} and DEAM~\citep{ghazarian-etal-2022-deam}, two state-of-the-art dialogue-level metrics. Detailed metric descriptions are outlined in Appendix~\ref{sec:sota-metrics}.


\subsection{Implementation Details}

The thresholds for the QA relevance strategy ($\tau^{rel}_{low}$, $\tau^{rel}_{high}$), the contradiction scoring strategy ($\tau^{contra}_{low}$, $\tau^{contra}_{high}$),  and the entailment scoring strategy ($\tau^{entail}_{low}$, $\tau^{entail}_{high}$) are heuristically set to (0.85, 0.99), (0.20, 0.40), (0.01, 0.10) respectively. These thresholds ensure that there are enough data instances within both the positive and negative class. 

Each experiment is repeated five times with different random seeds. Since we have prepared two training datasets, there are 5$\times$2$=$10 variants for $M^q$, $\text{FineD-Eval}_{en}$, and $\text{FineD-Eval}_{mu}$ respectively. In \S\ref{sec:exp-analysis}, we report the average Spearman correlation scores across the 10 variants. Additional details associated with the training process, such as hyperparameters, model selection criteria, etc. are included in Appendix~\ref{sec:reproducibility}. 

\begin{table*}[!ht]	
\centering
\begin{adjustbox}{max width=0.8\linewidth}
\begin{tabular}{@{}c|c|cccccccc|c@{}}
\toprule
\textbf{Groups} & \textbf{Metrics}  & \textbf{Coh} & \textbf{Und} & \textbf{Fle} & \textbf{Lik} & \textbf{Inf} & \textbf{Top} & \textbf{Div} & \textbf{Ove} & \textbf{Average}
\\ \midrule
\multirow{4}{*}{Turn} & USL-H & 19.50 & \textit{14.66} & 18.98 & 31.00 & 35.39 & 31.86 & 20.70 & 24.10 & 23.27 \\
& MAUDE & -22.37 & -28.12 & -28.18 & -33.12 & -32.76 & -25.50 & -19.67   & -28.05 & -27.22 \\
& MDD-Eval & 27.62 & 23.43  & \textit{8.35} & \textit{11.87}  & \textit{6.86} & \textit{-0.61} & \textit{-6.83} & 13.10 & 10.47 \\
& D-score & 31.15 & 31.14 & 32.77 & 27.04  & 23.82 & 22.17 & 20.83 & 37.58 & 28.31 \\ \midrule
\multirow{2}{*}{Dialogue} & DynaEval & 42.29  & 36.06  & 38.91 & 39.78 & 39.61  & 43.94 & 33.16 & 48.18 & 40.24 \\
& DEAM & 46.82 & 46.68  & 52.19 & 50.49  & 59.20 & 61.90 & 59.20 & 54.72 & 53.90 \\ \midrule
\multirow{3}{*}{Sub-metrics} & $M^{\text{Coh}}$ & 52.86 & 52.35 & 43.87 & 47.71 & 42.84 & 40.54 & 36.43 & 53.02 & 46.20 \\
& $M^{\text{Lik}}$ & 42.91 & 42.15 & 37.08 & 52.23 & 49.89 & 41.36 & 36.52 & 48.83 & 43.87 \\
& $M^{\text{Top}}$ & 23.25 & 25.87 & 36.04 & 36.93 & 46.63 & 56.53 & 53.38 & 36.31 & 39.37 \\ \midrule
\multirow{5}{*}{Combined} & $M^{\text{Coh}}+M^{\text{Lik}}$ & 57.61 & 57.13 & 48.77 & 61.30 & 57.20 & 49.94 & 44.29 & 61.35 & 54.70 \\
& $M^{\text{Coh}}+M^{\text{Top}}$ & 53.11 & 54.99 & 51.36 & 54.75 & 55.67 & 58.66 & 54.02  & 59.30 & 55.23 \\
& $M^{\text{Lik}}+M^{\text{Top}}$ & 45.43 & 46.87 & 44.78 & 57.35  & 59.20 & 56.58 & 50.71 & 55.10 & 52.00 \\
& $\text{FineD-Eval}_{en}$ & \textbf{58.30} & \textbf{59.49} & 53.74 & 64.75 & 64.17 & 61.23 & 55.09 & 65.47 & 60.28 \\
& $\text{FineD-Eval}_{mu}$ & 57.66 & 57.37 & \textbf{55.94} & \textbf{64.91} & \textbf{66.84} & \textbf{66.22} & \textbf{59.59} & \textbf{66.15} & \textbf{61.84} \\
\bottomrule
\end{tabular}
\end{adjustbox}
\caption{Spearman correlations (\%) of different metrics on FED. Coh, Und, Fle, Lik, Inf, Top, Div, and Ove denote coherence, understanding, flexibility, likability, informativeness, topic depth, diversity, and overall impression respectively. The scores w.r.t. the best performing metric for each quality dimension are highlighted in bold. Statistically insignificant scores ($p > 0.05$) are italicized.}
\label{tab:fed-correlation}
\end{table*}

\begin{table}[!ht]	
\centering
\begin{adjustbox}{max width=.9\linewidth}
\begin{tabular}{@{}c|cccccc@{}}
\toprule
 & USL-H & DEAM & $M^{\text{Coh}}$ & $\text{FineD-Eval}_{en}$ & $\text{FineD-Eval}_{mu}$ \\
\toprule 
Coh & 19.86 & 18.29 & \textbf{22.04} & 21.72 & 21.02 \\
Und & 17.82 & 18.89 & \textbf{19.89} & 19.43  & 19.02 \\
Fle & 18.31 & 18.49 & \textbf{20.57} & 19.95 & 19.32 \\ 
Lik & 18.62 & 16.82 & 20.76 & \textbf{21.82} & 21.60 \\
Inf & 15.76 & 14.17 & 15.34 & 17.64 & \textbf{18.14} \\
Top & 18.76 & 15.61 & 17.61 & 19.95  & \textbf{20.35} \\
Div & 12.95 & \textbf{16.92} & 12.87 & 14.36 & 14.38 \\
Ove & 19.77 & 19.37 & 22.89 & 22.94 & \textbf{23.01} \\ \midrule
Average & 17.73 & 17.32 & 19.00 & \textbf{19.73} & 19.60 \\
\bottomrule
\end{tabular}
\end{adjustbox}
\caption{Spearman correlations (\%) of different metrics on DSTC9-Interactive. USL-H  and DEAM are the best turn-level and dialogue-level baselines respectively. $M^{\text{Coh}}$ is the best performing sub-metric. Full results can be found at Table~\ref{tab:full-dstc9-correlation}.}
\label{tab:dstc9-correlation}
\end{table}

\begin{table}[!ht]	
\centering
\begin{adjustbox}{max width=.9\linewidth}
\begin{tabular}{@{}c|ccccc@{}}
\toprule
 & D-score & DEAM & $M^{\text{Coh}}$ &  $\text{FineD-Eval}_{en}$ & $\text{FineD-Eval}_{mu}$ \\
\toprule 
Interest & 11.50 & 7.02 & 17.69 & \textbf{19.72} & 19.31 \\
Sensible & 22.33 & 13.92 & \textbf{25.25} & 22.65 & 20.23 \\
Humanness & 13.16 & 8.27 & \textbf{20.31} & 19.15 & 16.84 \\
Enjoyment & 15.07 & 9.21 & 19.52 & \textbf{20.30} & 18.98 \\
Listening & 20.38 & 13.67 & \textbf{29.74} & 22.52 & 20.61 \\
Avoid Rep & 10.58 & 17.45 & \textbf{21.44} & 17.60 & \textbf{17.98} \\ 
Fluency & 20.47 & 17.85 & \textbf{22.29} & 19.75 & 19.35 \\
\midrule
Average & 16.21 & 12.49 & \textbf{22.32} & 20.24 & 19.04 \\
\bottomrule
\end{tabular}
\end{adjustbox}
\caption{Spearman correlations (\%) on Persona-Eval. "Avoid Rep" denotes avoid repetition. D-score and DEAM are the best turn-level and dialogue-level baselines respectively. $M^{\text{Coh}}$ is the best performing sub-metric. Full results can be found at Table~\ref{tab:full-persona-eval-correlation}.}
\label{tab:persona-eval-correlation}
\end{table}

\section{Experiments \& Analysis}
\label{sec:exp-analysis}

In this section, we conduct the main analysis based on the following research questions (RQ): (1) Are dialogue-level metrics better than turn-level metrics for multi-turn dialogue evaluation? (2) Does our proposed sub-metrics correlate well with human evaluation for their target dimensions? (3) Does combining different sub-metrics help achieve better correlations for different dimensions? (4) Does FineD-Eval offer more interpretable results? (5) How reliable are negative samples constructed with sampling strategies in \S\ref{subsec:submetrics}? Additional analyses are presented in Appendix~\ref{sec:additional-analysis}

\noindent \textbf{RQ 1.}\ First, we can observe in Table~\ref{tab:fed-correlation} that all dialogue-level metrics perform significantly better than the turn-level metrics across different quality dimensions, this observation is inline with conclusions from previous works~\citep{yeh-etal-2021-comprehensive,zhang-etal-2021-dynaeval}. However, USL-H and D-score outperform DEAM on On DSTC9-Interactive (Table~\ref{tab:dstc9-correlation}) and Persona-Eval (Table~\ref{tab:persona-eval-correlation}) respectively. The good performance of USL-H and D-score may be attributed to that both metrics are an ensemble of multiple sub-metrics whereas DEAM is a single-model metric. This supports our claim that combining fine-grained metrics yield a holistic one that achieve better correlation with human judgment. Nevertheless, $\text{FineD-Eval}_{en}$ and $\text{FineD-Eval}_{mu}$, two dialogue-level metrics, outperform the turn-level metrics across all the dialogue-level benchmarks. We can conclude that in general, dialogue-level metrics perform better than turn-level metrics for multi-turn dialogue evaluation. 

\bigskip
\noindent \textbf{RQ 2.}\ In the sub-metrics section of Table~\ref{tab:fed-correlation}, we present the result of each dimensions-specific sub-metric on FED. We can observe that for coherence, understanding, and flexibility, $M^{\text{Coh}}$ achieves the best performance in the sub-metrics group with 52.86\%, 52.35\%, and 47.71\% Spearman correlations respectively.  $M^{\text{Lik}}$ achieves the best performance in likability and informativeness with spearman correlations of 52.23\% and 49.89\% respectively. For topic depth and diversity, $M^{\text{Top}}$ performs the best among the three sub-metrics. The empirical results meet our expectation that the three sub-metrics target dimension groups 1, 2, and 3 in Table~\ref{tab:dim-categorization} respectively. For the coherence dimension, $M^{\text{Coh}}$ outperforms DynaEval and DEAM, which are also designed for evaluating dialogue-level coherence. Moreover, $M^{\text{Coh}}$ performs exceptionally well on DSTC9-Interactive (Table~\ref{tab:dstc9-correlation}) and Persona-Eval (Table~\ref{tab:persona-eval-correlation}) and significantly outperforms the turn-level and dialogue-level baselines on both benchmarks. This showcases the advantage of our utterance shuffling and QA relevance scoring strategies for coherence modeling. 

\bigskip
\noindent \textbf{RQ 3.} We can observe in Table~\ref{tab:fed-correlation} that combining different sub-metrics generally performs better than individual sub-metrics for various fine-grained dimensions. For example, $M^{\text{Coh}}+M^{\text{Lik}}$ outperforms $M^{\text{Coh}}$ for the coherence, understanding, and flexibility dimensions. It also outperforms $M^{\text{Lik}}$ for the informativeness and likability dimensions. Furthermore, metrics in the combined group significantly outperform the sub-metrics as well as various baselines for the overall impression dimension. The observations support our claim in the introduction that combining sub-metrics helps achieve better correlations for different quality dimensions. 

In addition, $\text{FineD-Eval}_{en}$ and $\text{FineD-Eval}_{mu}$ achieve a remarkable Spearman correlation of 65.47\% and 66.15\% respectively for the overall dimension on FED. Both outperforms state-of-the-art metrics as well as individual sub-metrics by a large margin. Such an observation showcases the effectiveness of metric ensemble and multitask learning for a holistic automatic dialogue-level evaluation. 

Similar observations can be made w.r.t. DSTC9-Interactive (Table~\ref{tab:dstc9-correlation} and Table~\ref{tab:full-dstc9-correlation}). Overall, $\text{FineD-Eval}_{en}$ and $\text{FineD-Eval}_{mu}$ achieve state-of-the-art performance on FED, DSTC9-Interactive, and Persona-Eval. Compared to the existing state-of-the-art metrics on the three benchmarks: DEAM, USL-H, and D-Score, $\text{FineD-Eval}_{en}$ achieves around 16\% relative improvement on average while $\text{FineD-Eval}_{mu}$ attains around 14\% relative improvement on average.

However, on Persona-Eval, the single metric, $M^{\text{Coh}}$ performs better than $\text{FineD-Eval}_{en}$ and $\text{FineD-Eval}_{mu}$ for all fine-grained dimensions except interestingness and enjoyment. A possible reason is that all the dimensions except interestingness and enjoyment are mostly related to coherence, Sub-metrics targeting likability and topic depth may be irrelevant for evaluating those dimensions. Incorporating scores from irrelevant sub-metrics may reduce the correlation with the corresponding human scores.




\begin{table}[ht!]
\small
\centering
\begin{tabular}{rp{5cm}}
\toprule
\multicolumn{2}{c}{{\bf Human-Bot Dialogue For Analysis}} \\
\midrule
\textbf{Human}: & \textit{Hi!} \\
\textbf{Bot}: & \textit{Hi there.} \\
\textbf{Human}: & \textit{What's your favorite food?} \\
\textbf{Bot}: & \textit{I require only kebabs.} \\
\textbf{Human}: & \textit{Why's that?} \\
\textbf{Bot}: & \textit{Because my robot machinery breaks down, the meat in kebabs to give me energy, in a similar way that you need food to survive.} \\
\textbf{Human}: & \textit{Why are kebabs special?} \\
\textbf{Bot}: & \textit{That's just the way it is.} \\
\textbf{Human}: & \textit{Who are you?} \\
\textbf{Bot}: & \textit{I am System. I want to be your friend.} \\
\textbf{Human}: & \textit{Why?} \\
\textbf{Bot}: & \textit{Because it's good to have many friends.} \\
\textbf{Human}: & \textit{How many is too many, though} \\
\textbf{Bot}: & \textit{Oh. More than a few.} \\
\bottomrule
\end{tabular}
\caption{A human-chatbot dialogue from FED. It receives 0.8, 0.3, 0.2, and 0.4 human ratings (out of 1.0) for coherence, likability, topic depth, and overall impression respectively. \label{tab:interpretable-sample}}
\end{table}



\bigskip
\noindent \textbf{RQ 4.}  Table~\ref{tab:interpretable-sample} is a case study to demonstrate that FineD-Eval provides more interpretable results than existing dialogue-level metrics. DEAM and DynaEval assign a metric score of 0.9935 and 0.2325 to the dialogue respectively. Both metrics only partially capture the dialogue quality. The DEAM score reflects the degree of coherence while the DynaEval score reflects the overall quality. However, even though the dialogue is coherent, human judges do not like the chatbot (0.3 likability rating) and the topics discussed in the dialogue is also not in depth (0.2 topic depth rating). These aspects are not captured by DEAM nor DynaEval. On the contrary, either $\text{FineD-Eval}_{en}$ or $\text{FineD-Eval}_{mu}$ can assign fine-grained scores that capture these aspects. The $\text{FineD-Eval}_{en}$ metric scores for coherence, likability, topic depth, and overall impression are 0.6123, 0.1865, 0.0632, and 0.2874 respectively. In this sense, FineD-Eval variants are more interpretable than existing metrics, because it helps dialogue researchers know exactly which dialogue aspect they should improve upon. 

\bigskip
\noindent\textbf{RQ 5.}  Table~\ref{tab:negative-dialog-samples} presents example human-human dialogues that are considered as negative samples. We examine them one by one to validate the reliability of our sampling strategies. First, in the incoherent dialogue obtained by the QA relevance strategy, we can observe that when speaker B asks "isn't it" and expect speaker A to acknowledge, A instead replies by "One last question.", which disrupts the coherent flow of the dialogue to a certain extent. The predicted QA relevance score of the exchange is 0.390, which suggests poor coherence. Speaker B in the second dialogue displays an uncooperative personality. Though the conversation is coherent, if a chatbot displays such a personality, user disengagement may happen. The contradiction score w.r.t. the dialogue example is 0.741, which suggests dislikability. In the third dialogue, we can observe that the utterances are short and do not contain much meaningful content. The predicted entailment score of the dialogue is 0.599, which indicates a lack of topic depth. Lastly, none of the utterance in the last dialogue example contains positive sentiment. The entire dialogue looks uninteresting. Overall, the qualitative examples support that our proposed strategies are reliable in generating negative samples.

\begin{table}[ht!]
\small
\centering
\begin{tabular}{rp{6cm}}
\toprule

\multicolumn{2}{c}{{\bf QA Relevance (Incoherence)}} \\
\midrule
\textbf{A}: & \textit{Oh , they're both so beautiful . Let me have this one , I think .} \\
\textbf{B}: & \textit{That one truly is a beautiful piece of work , isn't it ?} \\
\textbf{A}: & \textit{One last question .} \\
\textbf{B}: & \textit{Oh , no . Everything we sell here is ' as is ' .} \\
\midrule

\multicolumn{2}{c}{{\bf Contradiction Scoring (Dislikability)}} \\
\midrule
\textbf{A}: & \textit{We have a special on these skirts this week . Would you like to try one on ?} \\
\textbf{B}: & \textit{No , thank you . I don't need any skirts.} \\
\textbf{A}: & \textit{How about a blouse ? This one here is the latest fashion } \\
\textbf{B}: & \textit{No , thank you .} \\
\midrule

\multicolumn{2}{c}{{\bf Entailment Scoring (Dullness)}} \\
\midrule
\textbf{A}: & \textit{All right , so I'll see you then .} \\
\textbf{B}: & \textit{I'll call you later .} \\
\textbf{A}: & \textit{Okay , I'll talk to you later then .} \\
\textbf{B}: & \textit{See you later .} \\
\textbf{A}: & \textit{Bye .} \\
\midrule

\multicolumn{2}{c}{{\bf Number of Positive Utterances (Dislikability)}} \\
\midrule
\textbf{A}: & \textit{Is it okay to have a day off next week ?} \\
\textbf{B}: & \textit{Why ? What's the problem ?} \\
\textbf{A}: & \textit{I need to go to the dentist .} \\
\textbf{B}: & \textit{Okay , I'll get Bob to cover you .} \\
\bottomrule
\end{tabular}
\caption{Negative human-human dialogue examples obtained with QA relevance, contradiction scoring, entailment scoring, and sentiment strategies. \label{tab:negative-dialog-samples}}
\end{table}

\section{Related Work}
\label{sec:related}

Evaluation is a long-lasting problem in dialogue system research~\citep{deriu2021survey,yeh-etal-2021-comprehensive,mehri-etal-2022-nsf,smith-etal-2022-human}. In open-domain dialogue evaluation,~\citet{liu-etal-2016-evaluate} shows that commonly-adopted metrics, such as BLEU~\citep{papineni-etal-2002-bleu}, can be misleading due to their poor correlations with human judgment. Recently, interests in automatic evaluation of open-domain dialogue systems have intensified with the introduction of reference-free model-based evaluation paradigm\footnote{See~\citet{yeh-etal-2021-comprehensive} for a comprehensive list of recent evaluation metrics.}. Most of them focus on turn-level response quality~\citep{tao2018ruber,ghazarian-etal-2019-better,huang-etal-2020-grade,sai-etal-2020-improving,sinha-etal-2020-learning,zhang2021mdd}. Despite their promising correlations with human evaluation, such metrics are insufficient for dialogue-level assessment. Our FineD-Eval targets dialogue-level evaluation specifically.

In addition, existing works on model-based dialogue-level metrics~\citep{li-etal-2021-conversations,zhang-etal-2021-dynaeval,ghazarian-etal-2022-deam,zhao-etal-2022-floweval} focus very much on a single quality dimension. On the contrary, FineD-Eval is capable of multi-dimensional evaluation and it can provide more fine-grained and interpretable scores. 

The idea of decomposing overall dialogue quality into fine-grained dimensions has been explored in prior works~\citep{mehri-eskenazi-2020-usr,phy-etal-2020-deconstruct,pang-etal-2020-towards,zhang-etal-2021-dscore} for turn-level evaluation. However, its application on dialogue-level evaluation is under-explored, our work serves to bridge this gap.

\section{Conclusion}
\label{sec:concl}
In this paper, we propose FineD-Eval, a multi-dimensional dialogue-level evaluation metric. FineD-Eval consists of three specialized sub-metrics, which targets three fine-grained dialogue quality respectively, including coherence, likability, and topic depth. Each specialized sub-metric is trained with a pairwise ranking objective on dialogue pairs that are curated according to the corresponding dimension-specific strategies. Two variants of FineD-Eval are proposed to combine the sub-metrics into a holistic metric. One variant is based on metric ensemble and the other is based on multitask learning. We have empirically demonstrated that FineD-Eval strongly correlate with human evaluation for different dialogue quality dimensions as well as exhibits strong generalization across different evaluation datasets.

\section*{Limitations}
\label{sec:limitation}
We have identified two limitations that need to be addressed in future work. 

First, we can observe in Table~\ref{tab:dstc9-correlation} that the correlation scores of all the dialogue-level metrics including $\text{FineD-Eval}_{en}$ and $\text{FineD-Eval}_{mu}$ are much lower than those in Table~\ref{tab:fed-correlation}. There are two major reasons. The first reason is due to longer dialogues in DSTC9-Interactive than in FED (28.13 vs 12.72 utterances per dialogue). Existing metrics do not have effective mechanism to handle long dialogues. They often adopt BERT-based language models~\citep{devlin-etal-2019-bert,liu2019roberta} as the text encoders. As a result, longer dialogues are truncated to satisfy the input length and GPU memory constraints. Some information that is beneficial for dialogue-level evaluation may be lost due to truncation. In future, we should explore more sophisticated text encoders to model long dialogues. In addition, FineD-Eval should also incorporate mechanisms to pinpoint the most relevant or important information to evaluation within long dialogues, such as a dialogue breakdown detection module. Another reason is that dialogues in DSTC9-Interactive contain much more noise than those in FED. Human judges find it difficult to evaluate the dialogues, resulting in low inter-annotator agreements w.r.t. different fine-grained dimensions. The inter-annotator agreements for different dimensions range between 0.56 and 0.58 in terms of Spearman correlations. On the contrary, the quality of FED dialogues is better and the inter-annotator agreements of most dimensions are above 0.8. Besides designing more robust metrics, future work should also explore developing more high-quality dialogue-level evaluation benchmarks. 


Second, as stated in \S\ref{sec:human-eval-analysis}, fine-grained quality dimensions, such as consistency, error recovery, and inquisitiveness are not covered by FineD-Eval. Hence, we do not report the performance of FineD-Eval on these dimensions in the main analysis. For completeness, we present the performance of FineD-Eval for the missing dimensions on the three benchmarks in Table~\ref{tab:other-dims-results}.  We can observe that the correlations of both $\text{FineD-Eval}_{en}$ and $\text{FineD-Eval}_{mu}$ for these three dimensions are not as high as those for the other dimensions, such as likability, topic depth, and coherence. The observation is expected as we do not have dedicated sub-metrics to model consistency, error recovery, and inquisitiveness. Hence, the dimensions missing from FineD-Eval are worth a thorough future study on their definitions, application scenarios, and metric designs.


\begin{table}[!ht]	
\centering
\begin{adjustbox}{max width=.9\linewidth}
\begin{tabular}{@{}c|cccc@{}}
\toprule
\multicolumn{5}{c}{FED} \\
\midrule
 & D-score & DEAM & $\text{FineD-Eval}_{en}$ & $\text{FineD-Eval}_{mu}$ \\
\midrule
Con & 20.89 & 30.99 & 47.85 & 44.58  \\
Inq & \textit{17.11} & 37.21 & 45.49 & 45.48 \\
Err & 22.66 & 39.54 & 51.17 & 50.89 \\
\midrule
\multicolumn{5}{c}{DSTC9-Interactive} \\
\midrule
 & USL-H & DEAM & $\text{FineD-Eval}_{en}$ & $\text{FineD-Eval}_{mu}$ \\
\midrule 
Con & 9.57 & 7.81 & 12.16 & 13.02  \\
Inq & 12.62 & 11.95 & 14.52 & 14.54 \\
Err & 15.10 & 15.31 & 15.97 & 15.75 \\
\midrule
\multicolumn{5}{c}{Persona-Eval} \\
\midrule
 & D-score & DEAM & $\text{FineD-Eval}_{en}$ & $\text{FineD-Eval}_{mu}$ \\
\midrule
Inq & 15.93 & 10.61 & 12.55 & 11.44 \\
\bottomrule
\end{tabular}
\end{adjustbox}
\caption{Additional results on the three benchmarks. "Con", "Inq", and "Err" denote "consistency", "inquisitiveness", and "error recovery" respectively. DEAM is the best dialogue-level baseline on all datasets. USL-H is the best turn-level baseline on DSTC9-Interactive while D-score is the best turn-level baseline on FED and Persona-Eval.}
\label{tab:other-dims-results}
\end{table}

\section*{Acknowledgement}
We would like to thank all the reviewers for their constructive comments. This work is supported by Science and Engineering Research Council, Agency of Science, Technology and Research (A*STAR), Singapore, through the National Robotics Program under Human-Robot Interaction Phase 1 (Grant No. 192 25 00054); Human Robot Collaborative AI under its AME Programmatic Funding Scheme (Project No. A18A2b0046); Robert Bosch (SEA) Pte Ltd under EDB’s Industrial Postgraduate Programme – II (EDB-IPP), project title: Applied Natural Language Processing; This work is also supported by the Internal Project Fund from Shenzhen Research Institute of Big Data under Grant T00120220002. The work leading to these results is also supported by project BEWORD (PID2021-126061OB-C43) funded by MCIN/AEI/10.13039/501100011033 and, as appropriate, by “ERDF A way of making Europe”, by the “European Union”, and by Programa Propio - Proyectos Semilla: Universidad Politécnica de Madrid (VSEMILLA22LFHE).



\appendix
\section{Dialogue Corpora}
\label{sec:detail-dataset}
The two dialogue corpora for constructing our training/validation datasets are outlined below. Their detailed statistics are presented in Table~\ref{tab:data-statistics}. Table~\ref{tab:train-statistics} shows the number of positive and negative dialogues that are constructed with each strategy (described in \S\ref{subsec:submetrics}) for different data splits.

\begin{table}[!ht]
\centering
\begin{adjustbox}{max width=\linewidth}
\begin{tabular}{l|cc}
\toprule
\textbf{DailyDialog} & \textbf{training} & \textbf{validation} \\ \midrule
\#dialogues & 11,118 & 1,000  \\
\#utterances & 87,170 & 8,069 \\
\#words & 1,186,046  & 108,933 \\
Avg. \#utterances per dialogue & 7.84 & 8.07 \\ 
Avg. \#words per dialogue & 106.68 & 108.93 \\ \midrule
\textbf{ConvAI2} & \textbf{training} & \textbf{validation} \\ \midrule
\#dialogues & 17,878 & 1000  \\
\#utterances & 253,698 & 15,566 \\
\#words & 3,024,032  & 189,374 \\
Avg. \#utterances per dialogue & 14.19 & 15.57 \\ 
Avg. \#words per dialogue & 169.15 & 189.37 \\ \bottomrule
\end{tabular}
\end{adjustbox}
\caption{Human-Human Dialogue Corpora Statistics}\label{tab:data-statistics}
\end{table}

\bigskip
\noindent\textbf{DailyDialog}~\citep{li-etal-2017-dailydialog} contains high-quality and human-written dialogues spanning 10 general chit-chat topics, including ordinary life, school, culture \& education, attitude \& emotion, relationship, tourism, health, work, politics, \& finance. The dialogues are mainly for information exchange among interlocutors and social bond enhancement. They are also annotated with turn-level dialogue act and emotion labels.

\begin{table*}[!ht]
\centering
\resizebox{\linewidth}{!}{
\begin{tabular}{l|cc|cc|c}
\toprule
& \multicolumn{2}{c|}{\textbf{Coherence}} & \multicolumn{2}{c|}{\textbf{Likability}} & \multicolumn{1}{c}{\textbf{Topic Depth}} \\
\cmidrule{2-6}
\textbf{Data Split} & \textbf{QA Relevance} & \textbf{Utterance Shuffle} & \textbf{Contradiction} & \textbf{\#Pos Utterances} & \textbf{Entailment} \\ \midrule
DailyDialog (Train) & 16,122 / 4,606 & 32,779 / 32,779  & 9,387 / 6,001 & 4,752 / 8,738 & 3,573 / 1,885 \\
DailyDialog (Dev)  & 1,456 / 390 & 2,847 / 2,847  & 951 / 593 & 377 / 722 & 104 / 534 \\
ConvAI2 (Train) & 38,551 / 10,479 & 83,665 / 83,665  & 20,380 / 25,009 &  12,870 / 10,591 & 3,573 / 1,885 \\
ConvAI2 (Dev) & 3,202 / 775 & 6,845 / 6,845  & 1,564 / 2,142 & 791 / 463 & 314 / 120 \\
\bottomrule
\end{tabular}
}
\caption{Training Data Statistics. "/" is used to separate the number of positive and negative dialogues. For each data split and each strategy, the maximum number of dialogue pairs is the number of positive dialogues multiply by the number of negative dialogues. Due to the large number of combinations, we only sample 100K and 10K pairs for training and validation respectively.}\label{tab:train-statistics}
\end{table*}


\bigskip
\noindent\textbf{ConvAI2}~\citep{dinan2020second} is an extended dataset of Persona-Chat~\citep{zhang-etal-2018-personalizing}. Dialogues in ConvAI2 are grounded by the personas of the interlocutors. Two interlocutors in a dialogue play the roles described by the corresponding personas. Each persona contains at least 5 role description sentences. Throughout the dialogue, the two interlocutors try to be engaging, to know each other, and to find their mutual interests. In total, there are 1155 possible personas for training. Topic shifts are common in ConvAI2 dialogues as the interlocutors are continually introducing new information about themselves during their interaction.  

\section{Evaluation Benchmarks}
\label{sec:detail-benchmark}
\noindent\textbf{FED}~\citep{mehri-eskenazi-2020-unsupervised} consists of 125 dialogues, among which 40 are collected between a human and the Meena chatbot~\citep{adiwardana2020towards}, 44 are collected between a human and the Mitsuku chatbot, and the remaining 41 are human-human dialogues. Each dialogue is annotated by five human judges for 11 different quality dimensions, including coherence, error recovery, consistency, diversity, topic depth, likability, understanding, flexibility, informativeness, inquisitiveness, and overall impression. The definition of each dimension is outlined in Table~\ref{tab:dialog-dims}. The ratings of all the dimensions are based on the 1-3 Likert scale except that consistency scores range from 0 to 1 and overall scores range from 1 to 5. The inter-annotator agreements for all dimensions are above 0.8 in terms of Spearman correlations except consistency (0.562), diversity (0.789), and inquisitiveness (0.769). 

\bigskip
\noindent\textbf{DSTC9-Interactive}~\citep{gunasekara-etal-2020-dstc9} consists of 2200 human-chatbot dialogues. All the dialogues are collected during the human evaluation of 10 dialogue systems participating in the DSTC9\footnote{The Ninth Dialog System Technology Challenge.} "Interactive Evaluation of Dialog" shared task. Each dialogue is annotated by three human judges for the same 11 quality dimensions in FED. The inter-annotator agreements for coherence, error recovery, consistency, diversity, topic depth, likability, understanding, informativeness, flexibility, inquisitiveness, and overall impression in terms of Spearman correlations are 0.573, 0.566, 0.585, 0.562, 0.566, 0.563, 0.577, 0.569, 0.566, 0.565, and 0.566 respectively. We can observe that the inter-annotator agreements of all dimensions in DSTC9-interactive are much lower than those in FED.

\begin{table*}[htbp!]
\centering
\begin{adjustbox}{max width=\textwidth}
  \begin{tabular}{l||l}
  \toprule
  \bf Dimension & \bf Definition\\
  \midrule
  \tt Coherence & Throughout the dialogue, is the system maintaining a good conversation flow? \\
  \tt Error Recovery & Throughout the dialogue, is the system able to recover from errors that it makes? \\
  \tt Consistency & Throughout the dialogue, is the system consistent in the information it provides? \\
  \tt Diversity & Throughout the dialogue, does the system provides a diverse range of responses? \\
  \tt Topic Depth & Throughout the dialogue, does the system discuss topics in depth? \\
  \tt Likability & Throughout the dialogue, does the system display a likeable personality? \\
  \tt Understanding & Throughout the dialogue, does the system understand the user?  \\
  \tt Informativeness & Throughout the dialogue, does the system provide unique and non-generic information? \\
  \tt Flexibility & Throughout the dialogue, is the system flexible and adaptable to the user and their interests? \\
  \tt Inquisitiveness & Throughout the dialogue, does the system actively ask the user questions? \\
  \tt Overall Impression & The overall quality and user satisfaction of the dialogue. \\ \bottomrule
  \end{tabular}
  \end{adjustbox}
  \caption{Definitions of the eleven dialogue quality dimensions of FED~\citep{mehri-eskenazi-2020-unsupervised} and DSTC9-Interactive~\citep{gunasekara-etal-2020-dstc9}. The definitions are adapted from~\citet{mehri-etal-2022-nsf}.}
  \label{tab:dialog-dims}
\end{table*}

\bigskip
\noindent\textbf{Persona-Eval}~\citep{see-etal-2019-makes} contains 3316 dialogues. The dialogues are collected from human judges interacting with 28 different model configurations plus human-human conversations. At the end of each interaction, the human judge evaluates the entire interaction for eight different quality dimensions: avoiding repetition, interestingness, sensibleness, fluency, listening, inquisitiveness, humanness and engagingness. The ratings of all dimensions are on a 1-4 Likert scale except avoiding repetition, which is on a 1-3 Likert scale. Each model configuration is assessed with more than 100 human judges. The definitions of the eight dimensions are listed in Table~\ref{tab:persona-eval-dims}.


\begin{table*}[htbp!]
\centering
\begin{adjustbox}{max width=0.8\textwidth}
  \begin{tabular}{l||l}
  \toprule
  \bf Dimension & \bf Definition\\
  \midrule
  \tt Avoiding Repetition & How repetitive was this user? \\
  \tt Interestingness &  How interesting or boring did you find this conversation?\\
  \tt Sensibleness & How often did this user say something which did not make sense? \\
  \tt Fluency & How naturally did this user speak English? \\
  \tt Listening & How much did the user seem to pay attention to what you said? \\
  \tt Inquisitiveness & How much did the user try to get to know you? \\
  \tt Humanness &  Do you think this user is a bot or a human? \\
  \tt Engagingness &  How much did you enjoy talking to this user? \\\bottomrule
  \end{tabular}
\end{adjustbox}
  \caption{Definitions of the eight dialogue quality dimensions of Persona-Eval. The definitions are adapted from~\citet{see-etal-2019-makes}.}
  \label{tab:persona-eval-dims}
\end{table*}


\section{Metrics}
\label{sec:sota-metrics}
\noindent\textbf{USL-H}~\citep{phy-etal-2020-deconstruct} stands for \textbf{U}nderstandability, \textbf{S}ensibleness, and \textbf{L}ikability in \textbf{H}ierarchy. It measures the overall quality of a dialogue response based on a configurable composite function of three scores, which correspond to the three quality dimensions respectively. Understandability refers to the naturalness of a response, and a BERT-base valid utterance prediction model (BERT-VUP) is trained to predict whether a response is syntactically well-formed or not. Sensibleness denotes the contextual relevance of a response. A BERT-base next utterance prediction model (BERT-NUP) is trained to assess sensibleness. Likability quantifies how likeable a response is for a particular task. Likability can be configured to adapt to the end evaluation task. In~\citet{phy-etal-2020-deconstruct}, specificity is applied as the proxy of likability, which is measured with a BERT-base mask language model (BERT-MLM). The USL-H metric is trained on DailyDialog.


\bigskip
\noindent\textbf{MAUDE}~\citep{sinha-etal-2020-learning} is a reference-free metric tailored for online dialogue evaluation. MAUDE leverages DistilBERT~\citep{sanh-etal-2019-distilbert} to extract latent representations of utterances and captures the temporal transitions that exist between them. The authors propose different data augmentation techniques to augment both the positive and negative responses. For positive response augmentation, back-translation and a sequence-to-sequence generative model are used to generate positive response variants. For negative response augmentation, word drop, word repeat, and word order shuffle are proposed to create syntactically negative responses. Random utterance selection is adopted to generate semantically negative responses. MAUDE is trained in a contrastive manner with noise contrastive estimation (NCE) loss on the ConvAI2 dataset.

\bigskip
\noindent\textbf{MDD-Eval}~\citep{zhang2021mdd} is a reference-free metric for evaluating response appropriateness. MDD-Eval specifically targets multi-domain turn-level evaluation. It relies on data augmentation techniques and a self-training setup for improving generalization across different dialogue domains. It has been shown to achieve state-of-the-art performance on six turn-level dialogue evaluation benchmarks. The training dataset of MDD-Eval, MDD-Data, is constructed based on DailyDialog, ConvAI2, TopicalChat~\citep{gopalakrishnan2019topical}, and EmpatheticDialogues~\citep{rashkin-etal-2019-towards}. 

\bigskip
\noindent\textbf{D-score}~\citep{zhang-etal-2021-dscore} is a holistic and reference-free dialogue evaluation framework based on multitask learning. It measures four aspects of a dialogue response: language fluency, context coherence, logical consistency and semantic appropriateness. Each aspect is evaluated with the corresponding scorer. All the scorers share a common RoBERTa-base~\citep{liu2019roberta} text encoder and are jointly learned in a self-supervised and multitask manner. D-score measures the overall quality of a response by taking an unweighted average of scores assigned by all four scorers. D-score is trained on the ConvAI2 dataset.

\bigskip
\noindent\textbf{DynaEval}~\citep{zhang-etal-2021-dynaeval} adopts a structured graph to model a dialogue. It explicitly captures the speaker-level and utterance-level dependency via contextualized representation and graph convolution network. DynaEval is trained in a contrastive manner to distinguish natural human-human dialogues from negative samples. The negative samples are constructed by applying two types of perturbations to the original human-human dialogues, namely random utterance replacement and speaker-level shuffling. Three variants of DynaEval are provided in~\citet{zhang-etal-2021-dynaeval}, which are trained on DailyDialog, EmpatheticDialogues, and ConvAI2 respectively.

\bigskip
\noindent\textbf{DEAM}~\citep{ghazarian-etal-2022-deam} is a dialogue-level coherence evaluation metric. Its backbone is a RoBERTa-large binary classification model. DEAM relies on abstract meaning representation (AMR) to generate semantically incoherent dialogues. With AMRs, four semantic-level negative strategies are proposed, including coreference inconsistency, irrelevancy, contradictions, and decrease engagement. DEAM is the current published state-of-the-art metric on FED and DSTC9-Interactive. DEAM is trained on data constructed from both ConvAI2 and TopicalChat.

\section{Reproducibility}
\label{sec:reproducibility}
All the experiments are conducted on a single Tesla V100 32GB GPU. The implementation is based on Pytorch~\citep{NEURIPS2019_bdbca288} and the Hugging Face Transformers library~\citep{wolf-etal-2020-transformers}. Since the sub-metrics of FineD-Eval are trained with a pairwise ranking task, we adopt accuracy to determine the model performance. The checkpoint with the best validation accuracy is chosen to perform the dialogue evaluation task. For training, we adopt AdamW optimizer~\citep{loshchilov2018decoupled} with a constant learning rate of 1e-5 and a mini-batch size of 32. The number of training epochs is set to 10. The model is evaluated every 2000 steps. If the validation accuracy does not improve for ten consecutive checkpoints, we stop the training process. On average, the training time for one epoch is 45 minutes. 

For $\text{FineD-Eval}_{mu}$, we choose the checkpoint with the best average validation accuracy across the three pairwise ranking tasks to perform dialogue evaluation. During training, the same hyperparameters as those of the sub-metrics are adopted. In additional, a mini-batch is formed with training instances that are uniformly drawn from the training sets of the three pairwise ranking tasks at run time. On average, the training time for one epoch is approximately 2.5 hours. 

Since we apply padding at the mini-batch level, all dialogues that contain more than 10 utterances are splitted into sub-dialogues, i.e., each sub-dialogue contains less than 10 utterances. The splitting procedure is to avoid too much padding in a mini-batch if a long dialogue is present. In this way, the GPU memory can be better utilized during training. Note that we only apply this splitting procedure during model training, not during the dialogue evaluation process. 

All the baselines are implemented with the repository provided by~\citet{yeh-etal-2021-comprehensive}\footnote{\url{https://github.com/exe1023/DialEvalMetrics}}. Since the implementations of MDD-Eval, D-score, and DEAM are not included in~\citet{yeh-etal-2021-comprehensive}, we adopt their respective open-source code and checkpoints.

\section{Additional Analysis}
\label{sec:additional-analysis}

\subsection{Proxy Metrics vs Sub-metrics}

As described in \S\ref{subsec:submetrics}, we have applied different pre-trained models to score the human-human dialogues for training data preparation. Different scores are applied to group the dialogues into pairs: (1) dialogue-level QA relevance score for the coherence sub-metric; (2) dialogue-level contradiction score for the likability sub-metric; (3) number of utterances with positive sentiments in a dialogue for the likability sub-metric; (4) dialogue-level entailment score for the topic depth sub-metric. One may argue that instead of training new metrics based on the prepared dialogue pairs, an alternative approach is to directly apply (1) - (4) as proxy indicators of the fine-grained dimensions. We can observe in Table~\ref{tab:compare-proxy-to-submetrics} that $M^{\text{Coh}}$ performs significantly better than (1) for the coherence dimension on FED. $M^{\text{Lik}}$ performs significantly better than (2) and (3) for the likability dimension while $M^{\text{Top}}$ outperforms (4) by a large margin for the topic depth dimension. This is because without training, implicit dialogue-level features essential for measuring coherence, likability, and topic depth may not be captured by the proxy metrics.

\begin{table}[!ht]	
\centering
\small
\begin{adjustbox}{max width=\linewidth}
\begin{tabular}{@{}c|ccc@{}}
\toprule
\textbf{Metrics}  & \textbf{Coherence} & \textbf{Likability} & \textbf{Topic Depth} \\ \midrule
QA Relevance & 30.17 & 26.60 & 24.38  \\
Contradiction & 40.25 & 37.82 & 37.26 \\
Entailment & \textit{12.13} & \textit{13.66} & 34.66  \\ 
\#Pos Utterances & 18.21 & \textit{10.95} & \textit{9.88} \\ \midrule
$M^{\text{Coh}}$ & \textbf{52.86} & 47.71 & 40.54 \\
$M^{\text{Lik}}$ & 42.91 & \textbf{52.23} & 41.36 \\
$M^{\text{Top}}$ & 23.25  & 36.93 & \textbf{56.53} \\
\bottomrule
\end{tabular}
\end{adjustbox}
\caption{Compare the four proxy scores to the proposed model-based sub-metrics of FineD-Eval. The results are Spearman correlations (\%) on FED. Best score of each column is highlighted in bold. Statistically insignificant ($p > 0.05$) scores are italicized.}
\label{tab:compare-proxy-to-submetrics}
\end{table}

\subsection{Case Study}
Table~\ref{tab:good-dialogue-sample} presents a high-quality dialogue sample from the FED benchmark. According to the overall impression human ratings (from high to low), it ranks first. $\text{FineD-Eval}_{en}$, $\text{FineD-Eval}_{mu}$, DynaEval, and DEAM rank it at the third, the 5th, the 14th, and the 21st positions respectively.

\begin{table}[ht!]
\small
\centering
\begin{tabular}{rp{5cm}}
\toprule
\multicolumn{2}{c}{{\bf Human-Bot Good Dialogue Example}} \\
\midrule
\textbf{Human}: & \textit{Hi!} \\
\textbf{Bot}: & \textit{Hello. How's your week coming along?} \\
\textbf{Human}: & \textit{It's great, thanks. I'm trying to learn how to make croissants.} \\
\textbf{Bot}: & \textit{Wow that's interesting. I have baked cookies, but croissants seem much more sophisticated. Did you make any progress?} \\
\textbf{Human}: & \textit{I've done them once or twice so far, but they haven't been flakey enough. I'm trying to figure out why. What kind of cookies have you made?} \\
\textbf{Bot}: & \textit{Mint chocolate chips. I think your croissants not being flakey. could have something to do with your oven's temperature.} \\
\textbf{Human}: & \textit{Ah, good thought, thanks!} \\
\textbf{Bot}: & \textit{Have you thought about melting some chocolate into your croissants? They don't have to be something unhealthy. For example, melted dark chocolate is good for the heart, and makes the resulting croissants taste much better.} \\
\bottomrule
\end{tabular}
\caption{A high-quality dialogue example from FED. Human judges score it at 3.0/3.0, 3.0/3.0, 3.0/3.0, 5.0/5.0 for coherence, likability, topic depth, and overall impression respectively.\label{tab:good-dialogue-sample}}
\end{table}


Table~\ref{tab:bad-dialogue-sample} presents a low-quality dialogue sample from FED. According to the overall impression human ratings (from high to low), it ranks at the 119th out of 125 dialogues. $\text{FineD-Eval}_{en}$, $\text{FineD-Eval}_{mu}$, DynaEval, and DEAM rank it at the 113th, the 120th, the 70th, and the 78th positions respectively. 

We can observe that in both cases, $\text{FineD-Eval}_{en}$ and $\text{FineD-Eval}_{mu}$ correlate strongly with human evaluation compared to existing state-of-the-art metrics, including DynaEval and DEAM. The examples also demonstrate that dialogues receiving high overall impression scores are generally good in terms of coherence, likability, and topic depth whereas those perceived as low-quality dialogues by human judges also receive low scores for coherence, likability, and topic depth. This supports our motivation on applying metric ensemble and multitask learning to build a holistic metric for dialogue-level evaluation.

\begin{table}[ht!]
\small
\centering
\begin{tabular}{rp{5cm}}
\toprule
\multicolumn{2}{c}{{\bf Human-Bot Bad Dialogue Example}} \\
\midrule
\textbf{Human}: & \textit{Hi!} \\
\textbf{Bot}: & \textit{Hello.} \\
\textbf{Human}: & \textit{who am I talking to?} \\
\textbf{Bot}: & \textit{I am System. I want to be your friend.} \\
\textbf{Human}: & \textit{define friendship} \\
\textbf{Bot}: & \textit{Being friends.} \\
\textbf{Human}: & \textit{what does that mean} \\
\textbf{Bot}: & \textit{Does "that" refer to "Being friends."?}
\\
\textbf{Human}: & \textit{yes} \\
\textbf{Bot}: & \textit{I don't want to talk about that now.} \\
\textbf{Human}: & \textit{do you have any friends} \\
\textbf{Bot}: & \textit{I have human and robot friends.  A lot of people on the Internet chat with me.} \\
\textbf{Human}: & \textit{chatting is not the same as being a friend} \\
\textbf{Bot}: & \textit{I like to make friends online.} \\
\textbf{Human}: & \textit{so what do you guys chat about ?} \\
\textbf{Bot}: & \textit{I think about computer programming a lot.} \\
\textbf{Human}: & \textit{any other topics ?} \\
\textbf{Bot}: & \textit{No. I think that's probably it.} \\
\textbf{Human}: & \textit{Just computer programming?} \\
\textbf{Bot}: & \textit{What interests you about computers?} \\
\textbf{Human}: & \textit{nothing} \\
\textbf{Bot}: & \textit{You are a Luddite.} \\
\textbf{Human}: & \textit{pardon?} \\
\textbf{Bot}: & \textit{You are excused.} \\
\textbf{Human}: & \textit{what's a Luddite ?} \\
\textbf{Bot}: & \textit{A Luddite believes that advances in technology are inherently bad.} \\
\textbf{Human}: & \textit{I never said that but there is more to life than computers} \\
\textbf{Bot}: & \textit{But you implied it. I believe there is, yes.} \\
\textbf{Human}: & \textit{you believe there is what?} \\
\textbf{Bot}: & \textit{There is.} \\
\textbf{Human}: & \textit{there is what ?} \\
\textbf{Bot}: & \textit{There is no sense in what you are asking.} \\
\bottomrule
\end{tabular}
\caption{A low-quality dialogue example from FED. Human judges score it at 1.5/3.0, 0.0/3.0, 0.9/3.0, 1.5/5.0 for coherence, likability, topic depth, and overall impression respectively.\label{tab:bad-dialogue-sample}}
\end{table}

\begin{table*}[!ht]	
\centering
\begin{adjustbox}{max width=0.9\linewidth}
\begin{tabular}{@{}c|c|cccccccc|c@{}}
\toprule
\textbf{Groups} & \textbf{Metrics}  & \textbf{Coh} & \textbf{Und} & \textbf{Fle} & \textbf{Lik} & \textbf{Inf} & \textbf{Top} &  \textbf{Div} & \textbf{Ove} & \textbf{Average}
\\ \midrule
\multirow{2}{*}{Coherence} & Utterance Shuffling & 39.41 & 37.05 & 32.21 &  29.13 & 29.02 & 26.08 & 26.35 & 37.44 & 32.09 \\
& QA Relevance & 49.65 & 47.07 & 42.04 & 46.81 & 39.37 & 36.15 & 32.84 & 48.82 & 42.84 \\ \midrule
\multirow{2}{*}{Likability} & Contradiction Scoring & 45.91 & 42.21 & 38.40 & 48.16  & 49.77 & 48.86 & 39.39 & 48.81 & 45.19 \\
& \#Pos Utterances & 36.37 & 34.92 & 26.55 & 41.10 & 37.00 & 24.16 & 20.69 & 37.99 & 32.35 \\ \midrule
\multirow{1}{*}{Topic Depth} & Entailment Scoring & 23.25 & 25.87 & 36.04 & 36.93 & 46.63 & 56.53 & 53.38 & 36.31 & 39.37 \\ \midrule
\multirow{3}{*}{Sub-metrics} &  $M^{\text{Coh}}$ & \textbf{52.86} & \textbf{52.35} & \textbf{43.87} & 47.71 & 42.84 & 40.54 & 36.43 & \textbf{53.02} & \textbf{46.20} \\
& $M^{\text{Lik}}$ & 42.91 & 42.15 & 37.08  & \textbf{52.23} & \textbf{49.89} & 41.36 & 36.52 & 48.83 & 43.87 \\
& $M^{\text{Top}}$ & 23.25 & 25.87 & 36.04  & 36.93 & 46.63 & \textbf{56.53} & \textbf{53.38} & 36.31 & 39.37 \\ \bottomrule
\end{tabular}
\end{adjustbox}
\caption{We present the Spearman correlations w.r.t. each sampling strategy and each sub-metric on FED. Coh, Und, Fle, Lik, Inf, Top, Div, and Ove denote coherence, understanding, flexibility, likability, informativeness, topic depth, diversity, and overall impression respectively. The scores w.r.t. the best performing metric for each quality dimension are highlighted in bold. All the scores are statistically significant.}
\label{tab:strategy-ablation}
\end{table*}

\subsection{Ablation Study}
As described in Section~\ref{subsec:submetrics}, we have five different sampling strategies for three fine-grained quality dimensions. In this section, we show the impact of each strategy on the metric performance in Table~\ref{tab:strategy-ablation}. It can be observed that all the sampling strategies work as expected. Metrics that adopt "utterance shuffling" or "QA relevance" strategies exhibit better correlations for coherence and understanding than for other fine-grained dimensions. Metric using "entailment scoring" strategy performs better for topic depth and diversity. The "contradiction scoring" and "\#utterances with positive emotions" strategies contribute the most to the likability and informativeness dimensions. $M^{\text{Coh}}$, which leverages both "utterance shuffling" and "QA relevance" outperforms metrics that rely on only one of the two strategies. Similarly, $M^{\text{Lik}}$, which combines the strength of both "contradiction scoring" and "\#utterances with positive emotions" strategies, performs the best for likability and informativeness. However, metric that leverages only the "contradiction scoring" strategy outweighs $M^{\text{Lik}}$ for other fine-grained dimensions, such as coherence and topic depth. This showcases that the "contradiction scoring" strategy can also contribute to the evaluation of these dimensions.
 
\subsection{Additional Results}

In Table~\ref{tab:full-dstc9-correlation} and Table~\ref{tab:full-persona-eval-correlation}, we show the full results of different metrics on the DSTC9-Interactive and Persona-Eval benchmarks respectively. We can observe that most of the baselines perform poorly, except USL-H, D-score, and DEAM. A possible reason is that these three metrics capture dialogue features from different perspectives rather than focusing only on single aspect. USL-H and D-score are an ensemble of multiple sub-metrics while DEAM relies on four different AMR-based dialogue-level perturbation strategies that help the model spot semantic errors including contradiction, irrelevancy, decreased engagement, and coreference inconsistency. 

Further, $M^{\text{Coh}}$ performs exceptionally well than $M^{\text{Lik}}$ and $M^{\text{Top}}$ across all the fine-grained dimensions On both DSTC9-Interactive and Persona-Eval. The reason may be that the annotations on these two datasets are biased. For FED, there are five annotators for each dialogue and the inter-annotator agreements are strong across different dimensions. Hence, the annotation quality is very high. On the contrary, for DSTC9-Interactive, there are only three annotators per dialogue and the inter-annotator agreements across different dimensions are moderate. For Persona-Eval, there is only one annotator per dialogue. Hence, the annotations on DSTC9-Interactive and Persona-Eval may be biased towards dialogue features that are associated with coherence. The QA relevance and utterance shuffling strategies used by $M^{\text{Coh}}$ better capture such features than the other strategies.

Moreover, on DSTC9-Interactive, the combined metric, $M^{\text{Coh}}+M^{\text{Lik}}$, performs the best. $\text{FineD-Eval}_{en}$ and $\text{FineD-Eval}_{mu}$ perform generally well across all the fine-grained dimensions. These observations further support our conclusion to RQ3 in \S\ref{sec:exp-analysis}.

Lastly, on Persona-Eval, the combined metrics do not necessarily perform better than $M^{\text{Coh}}$. $M^{\text{Coh}}+M^{\text{Lik}}$ outperforms $M^{\text{Coh}}$ for interestingness, making sense, humanness, and enjoyment, but not for listening, avoiding repetition and fluency. This may be because the $M^{\text{Lik}}$ captures dialogue features that are more associated with the first four aspects than with the listening, avoiding repetition and fluency. 

\begin{table*}[!ht]	
\centering
\begin{adjustbox}{max width=0.8\linewidth}
\begin{tabular}{@{}c|c|cccccccc|c@{}}
\toprule
\textbf{Groups} & \textbf{Metrics}  & \textbf{Coh} & \textbf{Und} & \textbf{Fle} & \textbf{Lik} & \textbf{Inf} & \textbf{Top} &  \textbf{Div} & \textbf{Ove} & \textbf{Average}
\\ \midrule
\multirow{4}{*}{Turn} & USL-H & 19.86 & 17.82 & 18.31 & 18.62 & 15.76 & 18.76 & 12.95 & 19.77 & 17.73 \\ 
& MAUDE & 13.21 & 10.70 & 12.63  & 11.75 & 6.52 & 6.56 & 7.96 & 12.52 & 10.23 \\
& MDD-Eval & 14.79 & 12.07 & 9.67 & 10.32 & 7.66 & 8.61 & 5.31 & 13.10 & 10.19 \\
& D-score & 20.13 & 17.08 & 18.61 & 18.03 & 13.18 & 16.07 & 12.86 & 20.72 & 17.09 \\ \midrule
\multirow{2}{*}{Dialogue} & DynaEval & 8.65 & 6.45 & 6.80 & 7.01 & \textit{1.70} & \textit{1.38} & \textit{0.51} & 5.90 & 4.80 \\
& DEAM & 18.29 & 18.89 & 18.49 & 16.82 & 14.17 & 15.61 & \textbf{16.92} & 19.37 & 17.32 \\ \midrule
\multirow{3}{*}{Sub-metrics} & $M^{\text{Coh}}$ & 22.04 & 19.89 & \textbf{20.57} & 20.76 & 15.34 & 17.61 & 12.87 & 22.89 & 19.00 \\
& $M^{\text{Lik}}$ & 16.20 & 13.64 & 12.96 & 16.18 & 12.23 & 14.82 & 9.83 & 16.82 & 14.09 \\
& $M^{\text{Top}}$ & 8.82 & 8.56 & 10.47 & 10.93 & 11.50 & 12.77 & 9.21 & 10.54 & 10.35 \\ \midrule
\multirow{5}{*}{Combined} & $M^{\text{Coh}}+M^{\text{Lik}}$ & \textbf{23.30} & \textbf{20.07} & 20.22 & \textbf{22.26} & 16.49 & 19.03 & 13.67 & \textbf{23.58} & \textbf{19.83} \\
& $M^{\text{Coh}}+M^{\text{Top}}$ & 19.75 & 18.21 & 19.48 & 19.89 & 16.71  & 18.79 & 13.82 & 21.25 & 18.49 \\
& $M^{\text{Lik}}+M^{\text{Top}}$ & 15.77 & 14.22 & 14.45 & 17.06 & 14.80 & 16.68 & 11.51  & 17.34 & 15.23 \\
& $\text{FineD-Eval}_{en}$ & 21.72 & 19.43 & 19.95 & 21.82 & 17.64  & 19.95 & 14.36 & 22.94 & 19.73 \\
& $\text{FineD-Eval}_{mu}$ & 21.02 & 19.02 & 19.32 & 21.60 & \textbf{18.14} & \textbf{20.35} & 14.38 & 23.01 & 19.60 \\
\bottomrule
\end{tabular}
\end{adjustbox}
\caption{Spearman correlations (\%) of different metrics on DSTC9-Interactive. Coh, Und, Fle, Lik, Inf, Top, Div, and Ove denote coherence, understanding, flexibility, likability, informativeness, topic depth, diversity, and overall impression respectively. The scores w.r.t. the best performing metric for each quality dimension are highlighted in bold. Statistically insignificant scores ($p > 0.05$) are italicized.}
\label{tab:full-dstc9-correlation}
\end{table*}

\begin{table*}[!ht]	
\centering
\begin{adjustbox}{max width=0.8\linewidth}
\begin{tabular}{@{}c|c|ccccccc|c@{}}
\toprule
\textbf{Groups}& \textbf{Metrics}  & \textbf{Int} & \textbf{Sen} & \textbf{Hum} & \textbf{Enj} & \textbf{Lis} & \textbf{Rep} & \textbf{Flu} & \textbf{Average} \\ \midrule
\multirow{3}{*}{Turn} & USL-H & 6.09 & -4.68 & \textit{-1.80} & \textit{2.60} & \textit{2.67} & \textit{-2.45} & -4.32 & \textit{-0.27} \\ 
& MDD-Eval & 9.73 & 23.60 & 17.45 & 12.78 & 21.34 & 4.28 & 18.53 & 15.39 \\
& MAUDE & 8.41 & 18.51 & 7.90  & 12.21 & 10.87 & 3.45 & 17.68 & 11.29 \\
& D-score & 11.50 & 22.33 & 13.16 & 15.07 & 20.38 & 10.58 & 20.47 & 16.21 \\ \midrule
\multirow{2}{*}{Dialogue} & DynaEval & 6.03 & 10.47 & 9.11 & 8.88 & 11.84 & 3.93 & 9.89 & 8.59 \\
& DEAM & 7.02 & 13.92 & 8.27 & 9.21 & 13.67 & 17.45 & 17.85 & 12.49 \\ \midrule
\multirow{3}{*}{Sub-metrics} & $M^{\text{Coh}}$ & 17.69 & 25.25 & 20.31 & 19.52 & \textbf{29.74} & \textbf{21.44} & \textbf{22.29} & \textbf{22.32} \\
& $M^{\text{Lik}}$ & 15.20 & 17.52 & 15.48 & 16.69 & 14.83 & 8.09 & 13.42 & 14.46 \\
& $M^{\text{Top}}$ & 7.88 & \textit{0.56} & \textit{2.10} & 5.64 & \textit{2.69} & 5.49 & 7.03 & 4.48  \\ \midrule
\multirow{5}{*}{Combined} & $M^{\text{Coh}}+M^{\text{Lik}}$ & \textbf{19.78} & \textbf{25.32} & \textbf{20.96} & \textbf{21.63} & 25.37 & 16.32 & 20.58 & 21.43 \\
& $M^{\text{Coh}}+M^{\text{Top}}$ & 15.57 & 18.92 & 15.44 & 15.83 & 21.73 & 20.03 & 18.47 & 18.00 \\
& $M^{\text{Lik}}+M^{\text{Top}}$ & 15.37 & 14.36 & 12.99 & 15.15 & 12.25 & 10.66 & 12.70 & 13.35 \\
& $\text{FineD-Eval}_{en}$ & 19.72 & 22.65 & 19.15 & 20.30 & 22.52 & 17.60 & 19.75 & 20.24 \\
& $\text{FineD-Eval}_{mu}$ & 19.31 & 20.23 & 16.84 & 18.98 & 20.61 & 17.98 & 19.35 & 19.04 \\
\bottomrule
\end{tabular}
\end{adjustbox}
\caption{Dialogue-level Spearman correlations (\%) on Persona-Eval. Int, Sen, Hum, Enj, Lis, Rep, and Flu denote interestingness, sensibleness, humanness, enjoyment, listening, avoid repetition, and fluency respectively. The scores w.r.t. the best performing metric for each quality dimension are highlighted in bold. Statistically insignificant scores ($p > 0.05$) are italicized.}
\label{tab:full-persona-eval-correlation}
\end{table*}

\end{document}